# Spatial Features for Multi-Font/Multi-Size Kannada Numerals and Vowels Recognition

[1]B.V. Dhandra, [2]Mallikarjun Hangarge, [1]Gururaj Mukarambi

[1]Department of P.G. Studies and Research in Computer Science Gulbarga University, Gulbarga
[2]Department of Computer Science, Karnatak Arts, Science and Commerce College Bidar
`dhandra_b_v@yahoo.co.in,mhangarge@yahoo.co.in,gmukarambi@gmail.com`

*Abstract-* This paper presents multi-font/multi-size Kannada numerals and vowels recognition based on spatial features. Directional spatial features viz stroke density, stroke length and the number of stokes in an image are employed as potential features to characterize the printed Kannada numerals and vowels. Based on these features 1100 numerals and 1400 vowels are classified with Multi-class Support Vector Machines (SVM). The proposed system achieves the recognition accuracy as 98.45% and 90.64% for numerals and vowels respectively.

*Keywords:* Spatial Features, OCR, SVM

## I. INTRODUCTION

Due to the advancement in the computer technology every organization is dictated to implement the automatic processing activities of the organization. Automatic reading of the vehicle registration numbers, postal zip codes for sorting the mails, ID numbers, processing of bank cheques etc., is the applications of numeral recognition system. These problems address the need of developing a system for numerals recognition. In this direction, many researchers have developed the numeral recognition systems by using various feature extraction methods such as statistical features, topological and geometrical features, global transformation and series expansion features like Hough Transform, Fourier Transform, Wavelets, Moments, etc. A survey on different feature extraction methods for character recognition is reported in [1]. A review on different pattern recognition methods is given in [2]. Khan et al. [3] have reported the recognition of printed characters of any font and size for Roman alphabet. Multi-font numeral recognition without thinning based on directional density of pixels is reported in [4]. Recognition of Farsi/Arabic using Chain Codes discussed in [5]. Hanmandlu et al. [6] have proposed a Fuzzy based approach for recognition of Multi-Font numerals. An overview of OCR research in Indian scripts is reported in [7]. In recent years, few approaches have been proposed in the literature for the recognition of Kannada characters/numerals.

Ashwin et al. [8] have formed the three basic Zones for the underlying character image. Each Zone is divided into a number of circular tracks and sectors. On pixels in each angular region is used as a feature. Support vector machine was employed for the classification of characters and achieved an accuracy of 86.11%. A modified region decomposition method and optimal depth decision tree for the recognition of Kannada characters was used by Nagabhushan et al. [9]. Sanjeev Kunte et al. [10] have developed an OCR system for the recognition of basic characters of printed Kannada text, which works for different font size and font style. Each image was characterized by using Hu's invariant and Zernike moments. They have achieved the recognition accuracy as 96.8% with neural network classifier. Various types of moments have been used in the literature to recognize the image patterns in number of applications. Various forms of moments like regular moments [11] ,Geometric[12,13] Legendre moments[14], Zernike/pseudo-Zernike moments [15,16], radial and rotational moments [18,17] are considered for character and numeral recognition. Dhandra et al. [19] have proposed Multi-Font and Multi-Size Kannada numerals recognition using structural features. Rajput et al. [20] have proposed Multi-Font printed Kannada numeral recognition using crack codes and Fourier descriptors. This motivated us to design a simple algorithm for printed Kannada numerals and vowels recognition that achieves the highest recognition accuracy with minimum number of features.

This paper is organized as follows: Section 2 contains the data collection and preprocessing of the images. Feature extraction procedure is discussed in Section 3.The experimental details and results obtained are presented in Section 4 and conclusion is the subject matter of Section 5.

## II. DATA SET AND PREPROCESSING

Kannada language is one of the fourth major south Indian languages spoken by about 50 million people. The Kannada script consists of 16 vowels, 36 consonants and 10 numerals. There is no standard database is available of Indian script characters. Therefore, we have created numerals and vowels using Nudi software with different font size varying from 16 to 50 points with different styles namely, Nudi Akshara-01, Nudi Akshara-02, Nudi Akshara-03, Nudi Akshara-04, Nudi Akshara-05, Nudi Akshara-06, and Nudi Akshara-07 and obtained its printouts. Further, printed pages were scanned through a flat bed HP Laser Jet M1522N scanner at 300 DPI and binarized using Global threshold method (i.e. Otus's method) and stored it in (.bmp) file format. The noise is removed by using Median filter. A sample dataset of printed

Kannada numerals and vowels with different font styles are presented in Fig.1 and Fig.2.

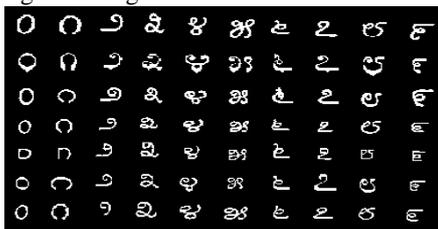

Figure 1: A sample dataset of printed Kannada numerals with different font styles

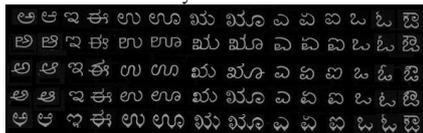

Figure 2: A sample dataset of printed Kannada vowels with different font styles

## III. FEATURE EXTRACION

The proposed feature extraction technique and its flow of execution is described below:

To perform basic morphological transformations, a line-structuring element (SE) is used. The length of structuring element is computed as threshold value multiplied by the height of the input image. The threshold values are experimentally fixed as 30%, 40%, 50%, 60%, 70% and 80% of the height of the input image. The variation in the recognition rate with different threshold values is shown in Fig. 4 (for Printed Kannada numerals).

Following operations are performed for feature extraction of an input image for its classification.

1. Perform directional opening of a input character image I by

$$\gamma_{(\theta,\mu)}(I) = \varepsilon_{(\theta,\mu)}[\delta_{(\theta,\mu)}(I)], \quad (1)$$

2. Compute the stroke length, which is defined as the number of pixels in a stroke as the measure of its length [21]. Average stroke length is defined as the average of the length of the individual strokes obtained in an image I. Thus, average stroke length ($\omega_\theta$) is given by of an image.

$$\omega_\theta = \frac{1}{n}\sum_{i=1}^{n} length(stroke_i) \quad (2)$$

for $\theta = 0^0, 45^0, 90^0$ and $135^0$, where $\omega_\theta$ is a feature vector of size 1x4, where n is number of strokes in an image.

3. Obtain the average stroke density ($\nu$), which is defined as the number of strokes per unit length (x-axis) of the input image I, which is computed by using the formula

$$\nu(\theta) = \frac{1}{n}\sum_{i=1}^{n}\left(\frac{n_i}{w(I)}\right) \quad (3)$$

Where w(I) is a width of the image.

for $\theta = 0^0, 45^0, 90^0$ and $135^0$, where the size of $\nu$ is 1x4. The sum of this row vector is used as average stroke density.

4. Compute the on-pixel ratio ($\eta$) using

$$\eta = \sum_{i=1}^{M}\sum_{j=1}^{N}\frac{f(i,j)}{size(I)} \quad (4)$$

In other words, on-pixel ratio is the ratio of on-pixels remaining after hole filling of the input image f to its size.

5. Compute the aspect ratio ($\beta$) using

$$\beta = \frac{width(f)}{height(f)} \quad (5)$$

That is, aspect ratio is the ratio of the width to the height of an input image [21].

6. **Eccentricity:** It is the length of major axis divided by the length of the minor axis of an input image f.

7. **Extent:** It is the proportion of the pixels in the bounding box that are also in the region. It can be computed as area (the number of pixels in a region) divided by the area of the bounding box.

8. **Directional profiles**: off pixels count of left, right, top and bottom of an input character is referred as directional profiles. A sample of four directional profile and directional stroke images of printed Kannada numerals and vowels are shown in Fig. 3 (a) and Fig. 3(b).

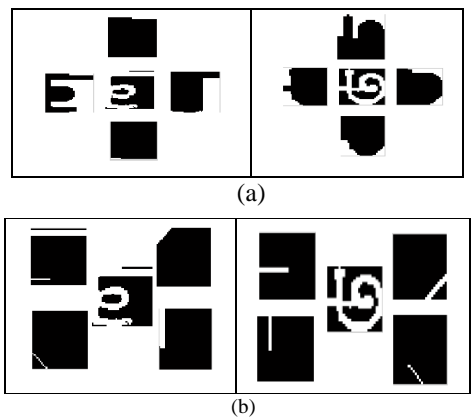

Figure 3: Printed Kannada numeral and vowel (a) Directional Profile (b) Directional Strokes

**Algorithm:** Recognition of printed Kannada numerals and vowels.
**Input:** Preprocessed Printed Kannada numeral and vowel image.

**Output:** Recognized numeral and vowel.
**Begin**
1. Compute the four directional openings of a character.
2. Compute stroke length using (2).
3. Compute the average stroke density using (3).
4. Compute on pixels ratio using (4).
5. Compute the aspect ratio, eccentricity, extent and directional profiles using equations (5), (6), (7) and (8) respectively.
6. Store a feature vector for further analysis.
7. Use SVM algorithm with sigma=0.6 to recognize printed Kannada numerals and vowels of Kannada script based on the feature vector obtained in step6.

**End**

## IV. EXPERIMENTAL RESULTS AND DISCUSSIONS

A sample data of 1100 printed Kannada numerals and 1400 Kannada vowels each of size varies from 16 to 50 points with 7 font styles are used for experimentation by considering 50% of the data for training and 50% for testing. The percentage of numerals recognition is 98.45% and vowels recognition is 90.64% were achieved with SVM classifier by fixing the threshold value of SE as 70% of the input image. Further, the recognition accuracy of 99.23% and 93.51% for Kannada numerals and vowels respectively were observed on three different font styles with the same threshold value. Details of the classification accuracies are presented in Table 1 and 2 for numerals and vowels respectively.

Table 1 Percentage of printed Kannada numerals recognition accuracy using SVM classifier with Sigma=0.6

| Training samples =550, Test samples =550 and Number of features = 13 | | | |
|---|---|---|---|
| Printed Kannada Numeral | No. of Sample Trained | No. of Sample Tested | Percentage of Recognition Accuracy |
| ೦ | 55 | 55 | 100.00 |
| ೧ | 55 | 55 | 96.61 |
| ೨ | 55 | 55 | 98.24 |
| ೩ | 55 | 55 | 100.00 |
| ೪ | 55 | 55 | 98.14 |
| ೫ | 55 | 55 | 96.15 |
| ೬ | 55 | 55 | 96.96 |
| ೭ | 55 | 55 | 100.00 |
| ೮ | 55 | 55 | 98.43 |
| ೯ | 55 | 55 | 100.00 |
| Average Percentage of Recognition accuracy = 98.45 | | | |

Table 2 Percentage of printed Kannada vowels recognition accuracy using SVM classifier with Sigma=0.6

| Training samples =700, Test samples =700 and Number of features = 13 | | | |
|---|---|---|---|
| Printed Kannada Vowel | No. of Sample Trained | No. of Sample Tested | Percentage of Recognition Accuracy |
| ಅ | 50 | 50 | 64.81 |
| ಆ | 50 | 50 | 75.47 |
| ಇ | 50 | 50 | 98.18 |
| ಈ | 50 | 50 | 94.33 |
| ಉ | 50 | 50 | 87.75 |
| ಊ | 50 | 50 | 98.11 |
| ಋ | 50 | 50 | 92.68 |
| ಌ | 50 | 50 | 97.87 |
| ಎ | 50 | 50 | 93.33 |
| ಏ | 50 | 50 | 92.30 |
| ಐ | 50 | 50 | 87.23 |
| ಒ | 50 | 50 | 95.00 |
| ಓ | 50 | 50 | 98.27 |
| ಔ | 50 | 50 | 93.61 |
| Average Percentage of Recognition accuracy = 90.64 | | | |

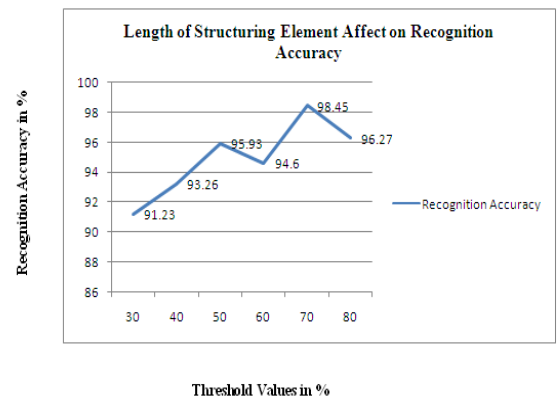

Figure 4: Graphical Representation of Length of Structuring Element Affect on Recognition Accuracy

**Comparative Analysis:**
The Table 3 and 4 shows the comparison results of existing methods with proposed method.

Table 3 Comparative results for printed Kannada vowel with other existing methods

| Name | Features and Classifiers | No. Features used | Recognition rate |
|---|---|---|---|
| Dinesh Acharya [26] | Topological Tree and knn | 48 | 92.32% |
| Proposed | Spatial and svm | 13 | 90.64% |

From Table 3, it is clear that proposed features are lesser than existing method features.

Table 4 Comparative results for printed Kannada numeral with other existing methods

| Name | Features and Classifiers | Recognition rate |
|---|---|---|
| Dhandra [27] | Modified Invariant Moment and NN | 98.92% |
| Proposed | Spatial K-NN | 98.45% |

## V. CONCLUSION

In this paper, a simple algorithm for Muli-font/Multi size printed Kannada numerals and vowels recognition is proposed based on the spatial features. The proposed method has shown quite encouraging performance with respect to printed Kannada numerals and vowels. This algorithm is independent of size normalization, thinning of the characters. However, it is dependent on the size and type of the structuring element used for feature extraction. An effort is on to design single OCR system for basic Kannada character set.

## ACKNOWELDGEMENT

This work is supported by UGC, New Delhi under Major Research Project grant in Science and Technology, (F.No-F33 - 64/2007 (SR) dated 28-02-2008). Authors are grateful to UGC for their financial support